\documentclass[runningheads]{llncs}
\usepackage{svg}
\usepackage{graphicx}
\usepackage[export]{adjustbox}
\usepackage{enumerate}
\usepackage{amssymb}
\usepackage[subrefformat=parens,labelformat=parens]{subfig}
\usepackage{amsmath}
\usepackage{multirow}
\usepackage{booktabs}
\usepackage{textgreek}
\usepackage{hhline}
\usepackage{array}
\usepackage{tabu}
\usepackage{verbatim} 
\usepackage{graphics}
\usepackage{color}
\usepackage{url}
\usepackage[colorlinks = true]{hyperref}
\usepackage{pifont}
\usepackage{makecell}

\newcommand{\ie}{\emph{i.e. }}
\newcommand{\etc}{\emph{etc. }}
\newcommand{\etal}{\emph{et al. }}
\newcommand{\cmark}{\ding{51}}
\newcommand{\xmark}{\ding{55}}
\begin{document}
\title{SelfDocSeg: A Self-Supervised vision-based Approach towards Document Segmentation}
\titlerunning{SelfDocSeg: A Self-Supervised Approach towards Document Segmentation}
\author{Subhajit Maity$^*$\inst{1}\orcidID{0000-0002-0735-8406} \and
Sanket Biswas$^*$\inst{2}\orcidID{0000-0001-6648-8270} \and
Siladittya Manna\inst{3}\orcidID{0000-0001-6364-8654} \and
Ayan Banerjee\inst{2}\orcidID{0000-0002-0269-2202} \and
Josep Lladós\inst{2}\orcidID{0000-0002-4533-4739} \and
Saumik Bhattacharya\inst{4}\orcidID{0000-0003-1273-7969}
\and
Umapada Pal\inst{3}\orcidID{0000-0002-5426-2618}}
% %
\authorrunning{S.Maity \etal}
\institute{Technology Innovation Hub, Indian Statistical Institute, Kolkata, India \\ 
              \email{subhajit\_t@isical.ac.in}\\
              \and
              Computer Vision Center, Computer Science Department, \\
              Universitat Autònoma de Barcelona, Spain \\
%              Fax: +123-45-678910\\
              \email{\{sbiswas, abanerjee, josep\}@cvc.uab.es} 
              %  \\
              \and
              CVPR Unit, Indian Statistical Institute, Kolkata, India \\ 
              \email{\{siladittya\_r, umapada\}@isical.ac.in}\\
              \and
              Department of Electronics and Electrical Communication Engineering,\\
              Indian Institute of Technology Kharagpur, India\\
              \email{saumik@ece.iitkgp.ac.in}}
\maketitle              % typeset the header of the contribution
\def\thefootnote{*}\footnotetext{These authors contributed equally to this work.}

\begin{abstract}
Document layout analysis is a known problem to the documents research community and has been vastly explored yielding a multitude of solutions ranging from text mining, and recognition to graph-based representation, visual feature extraction, \etc However, most of the existing works have ignored the crucial fact regarding the scarcity of labeled data. With growing internet connectivity to personal life, an enormous amount of documents had been available in the public domain and thus making data annotation a tedious task. We address this challenge using self-supervision and unlike, the few existing self-supervised document segmentation approaches which use text mining and textual labels, we use a complete vision-based approach in pre-training without any ground-truth label or its derivative. Instead, we generate pseudo-layouts from the document images to pre-train an image encoder to learn the document object representation and localization in a self-supervised framework before fine-tuning it with an object detection model. We show that our pipeline sets a new benchmark in this context and performs at par with the existing methods and the supervised counterparts, if not outperforms. The code is made publicly available at: \href{https://github.com/MaitySubhajit/SelfDocSeg}{github.com/MaitySubhajit/SelfDocSeg}

\end{abstract}
\section{Introduction}\label{sec:intro}
From the early days of computer vision and document understanding research, document layout analysis (DLA) had been a primitive problem and had been conquered with a multitude of strategies~\cite{binmakhashen2019document} from classical methods~\cite{agrawal2009voronoi++,marinai2005artificial,fang2011table} to state-of-the-art learning-based models~\cite{huang2022layoutlmv3,shen2021layoutparser,biswas2022docsegtr}. While intelligent document processing (IDP) has emerged as an essential step toward automatic document understanding, bearing the rise of convolutional neural networks (CNN) and sequence models, researchers have achieved near-perfect accuracy in the context of DLA with models having vast deployability and reliability. With a rapidly growing population, and the span of digital personal lives, documents are becoming more unconventional and business applications for IDP are encountering complex and never-before-seen layouts taking the already solved challenge a level up with the requirement of deep features exploration and exploitation. To meet that end, the state-of-the-art DLA strategies had improved a lot over time and still continue to do so, while remaining one of the most trendy topics for the research community~\cite{markewich2022segmentation}.

The challenge of document segmentation or DLA had been a research interest for nearly a decade and thus had been solved with classical heuristic rule-based methods~\cite{fang2011table,agrawal2009voronoi++} in the early days, while modern documents understanding community treats it as a document object detection (DOD) problem and is typically approached by a standard vision based object detector models~\cite{he2017mask,lin2017focal,redmon2016you}. With the breakthrough in sequence models and language models, researchers had been using the same along with object detection to solve the problem with better accuracy~\cite{huang2022layoutlmv3}. However, the research community has vastly ignored the fact that the growth in the number of unconventional documents raises the need for tedious annotation tasks to exploit the knowledge for better document understanding through the conventional supervised setting, and thus the self-supervised approaches toward the problem are highly relevant in this context.
\begin{figure*}[t]
\centering
\includegraphics[width=\linewidth]{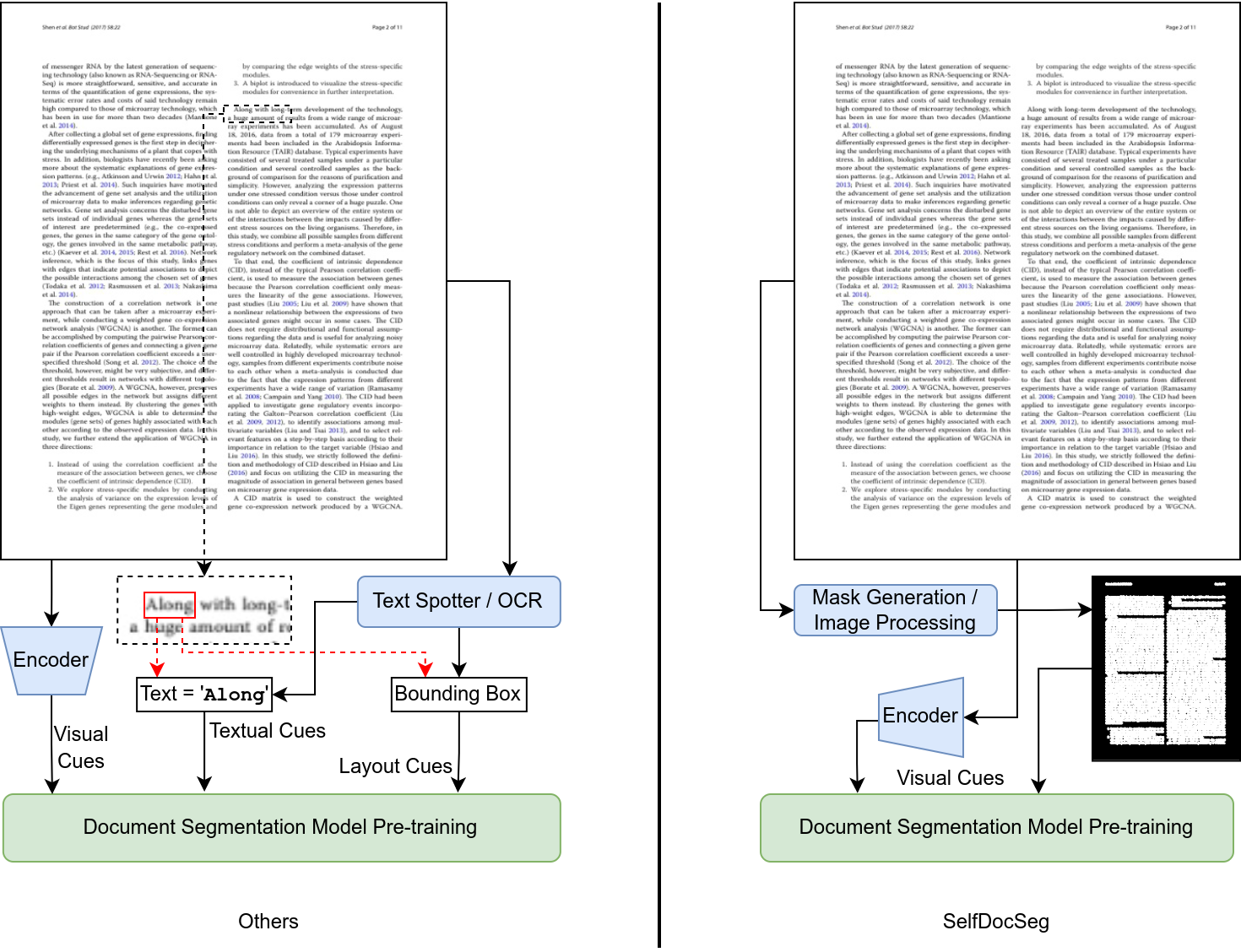}
%\vspace{-0.6cm}
\caption{
\textbf{Scope and Motivation}. {A basic methodological distinction between SelfDocSeg and existing approaches. While earlier works utilize information from visual, layout, and textual modalities for large-scale pre-training, we deal with visual cues only for boosting representation learning.}
}
\label{fig:teaser}
%\vspace{-0.4cm}
\end{figure*}
Deploying self-supervision in DOD is quite non-trivial as the task inherently cannot use the power of contrastive learning as the images contain multiple document objects of different classes and thus naively using each picture as its own class is not going to help. Apart from handling the class information, the encoder needs to consider object localization which cannot be realized through self-supervision without preliminary knowledge of the locations of layout objects. Moreover, it had been observed that although deep visual features extracted from document images are rich in information, they are difficult to train and are usually guided with learned textual~\cite{gu2021unidoc} or layout~\cite{biswas2022docsegtr} cues or both~\cite{huang2022layoutlmv3} from existing text spotter and detector models. And thus the efficacy of self-supervision for document segmentation faces a big question mark, which we address in this paper.

The core of our design consists of a self-supervised framework designed following 'Bootstrap Your Own Latent' (BYOL)~\cite{grill2020bootstrap} which actively tries to fulfill both the objectives of representation and localization. As self-supervision is employed for pre-training of the encoder, we do not have access to the ground-truth document object locations and categories. Thus we use classical image processing to approximate a rough physical layout mask for each document image and use the same as guidance for both document object localization and representation learning purpose. We use a backbone image encoder to obtain feature maps, followed by a mask pooling operation to extract encodings of all the possible relevant physical layout objects and train them in a self-supervised representation learning framework using negative cosine similarity. In parallel, we use a layout predictor module on the encoded feature maps,l and the module is tasked with a classification task of predicting if the salient features at every pixel of the feature map belong to a document object or not. This module is trained using focal loss~\cite{lin2017focal} with the supervision of the generated physical layout mask.

% The intuition behind our design is to devise the framework in a way it caters to object representation and localization simultaneously. To this end, the encoded feature map and mask pooling operation together draw out all the possible layout object representations simultaneously instead of one representation like BYOL~\cite{}. On the other hand, the layout predictor module learns to localize the regions of interest with the guidance of the layout mask. In a word, the model learns the physical layout representations and localization well enough and the logical layout objects detection becomes easier to distill in fine-tuning stage from learned physical layout objects representation and localization during pre-training.

The evidence had not been in favour of self-supervised vision-based approaches for DLA tasks as visual representations needed to be guided with learned textual and layout embeddings. However, we suggest that self-supervised visual representation learning can still be explored as visual features prove to be useful in the supervised settings for the task at hand. This had become the core motivating factor for our work as we try to explore and exploit the rich visual features extracted from the document images using the powerful CNN backbones. The intuition behind the proposed framework is relatively simple as we try to guide the learning of the backbone encoder using approximated visual layouts for both document object localization and representation learning. Unlike, the layout cues in the existing self-supervised strategies, we do not use any layout information from any pre-trained text recognition model. Instead, we devise the inherent visual information as a layout to guide the visual representation learning. A clear distinction in working principles between the proposed framework, SelfDocSeg, and the existing self-supervised strategies can be realized in Fig.~\ref{fig:teaser}.

\noindent
In summary, our contributions in this paper can be divided into three folds: \\
(a) A novel \textit{vision-based self-supervised framework}, specifically designed to pre-train an image encoder for DLA task; \\
(b) A \textit{pseudo physical layout guided strategy for self-supervision} in the region of interest localization for document segmentation; \\
(c) A \textit{data efficient pre-training strategy} to learn multiple document object representations simultaneously in the self-supervised setting.

\noindent
The rest of the paper is organized as follows. A comprehensive literature review is provided in Section~\ref{sec:lit}. The proposed methodology is described in Section~\ref{sec:method}.  In Section~\ref{sec:exp} we discuss the experiments and results. Finally, the conclusions are drawn in Section~\ref{sec:conclusion}.

\section{Related Work}\label{sec:lit}
\subsection{Self-Supervision for Visual Representation}
As the world of computer vision evolved it demanded undivided attention toward the exploration and exploitation of complex visual features to learn representations from images from a multitude of sources. Thus, data-centric machine-learning models with immense capability of feature extraction and correlation have emerged as modern-day solutions to the growing sophisticated requirements. However, with the huge amount of data required for modern network architectures, the need for data annotation has increased giving rise to the plethora of self-supervised strategies. MoCo~\cite{he2020momentum} provided the research community with the idea of weight update using the exponential moving averages and large memory banks in contrastive learning settings. SimCLR~\cite{pmlr-v119-chen20j} improved on it and introduced a large batch size as an alternative to the memory bank. DINO~\cite{caron2021emerging} introduced self-supervision in vision transformers~\cite{dosovitskiy2020image}. Following suit, MoCov2~\cite{chen2020improved}, SwAV~\cite{caron2020unsupervised} had achieved wonderful performance in the self-supervised paradigm. On the other hand, BYOL~\cite{grill2020bootstrap}, SimSiam~\cite{chen2021exploring} treat two crops from the same image as similar pairs instead of contrastive learning, while masked autoencoders~\cite{he2022masked} introduced a masking strategy to the age-old autoencoders for learning representation via reconstruction.

Among this plethora of self-supervised methods, self-supervised object detection and document segmentation remained vastly unexplored. Having significantly superior performances from supervised object detection methods like Mask RCNN~\cite{he2017mask}, Yolo~\cite{redmon2016you}, Retinanet~\cite{lin2017focal}, DETR~\cite{carion2020end} \etc, their self-supervised counterparts had been comparatively paler in diversity until recently. In the last few years, we have seen end-to-end self-supervised object detection models like UP-DETR~\cite{dai2021up}, DETReg~\cite{bar2022detreg} and backbone pre-training strategies like Self-EMD~\cite{liu2020self}, Odin~\cite{henaff2022object}. However, being closely related to object detection, instance-level document layout analysis has barely adopted self-supervision. Although there had been a few self-supervised document segmentation approaches, none of them had been explicitly using self-supervised visual representation or guidance. We address this lack of vision-based self-supervision for document understanding in this paper.

\subsection{Document Understanding}

Document Understanding (DU) has been reformulated in the current document analysis literature~\cite{borchmann2021due} as a landscape term covering different problems and tasks related to Document Intelligence systems which majorly includes Key Information Extraction~\cite{jaume2019funsd,park2019cord,stanislawek2021kleister}, Classification~\cite{harley2015evaluation}, Document Layout Analysis~\cite{zhong2019publaynet,hj2020dataset}, Question Answering~\cite{mathew2021docvqa,tito2022hierarchical}, and Machine Reading Comprehension~\cite{tanaka2021visualmrc} whenever they involve visually rich documents (VRDs) in contrast to plain texts or image-text pairs. Recent state-of-the-art DU systems majorly rely on large-scale pre-training to combine both visual and textual modalities as in ~\cite{appalaraju2021docformer,huang2022layoutlmv3,li2021selfdoc,gu2021unidoc,gemelli2023doc2graph} while methods like Donut~\cite{kim2021donut} and Dessurt~\cite{davis2023end} mostly rely on combining more effective visual features by synthetic generation techniques~\cite{biswas2021docsynth,yim2021synthtiger,coquenet2023dan,kang2021content} for learning important layout representation during document pre-training. In this work, we aim to identify a new direction toward boosting visual representation through a self-supervision strategy for document layout understanding.   

\subsection{Document Layout Analysis} DLA has evolved as a significant DU application, dedicated towards the optimization of storage and large-scale information workflows~\cite{binmakhashen2019document}. Since the advent of deep learning and CNN-based approaches~\cite{schreiber2017deepdesrt,shen2021layoutparser}, the segmentation of document layouts has been reformulated as DOD. With the launch of large-scaled annotated DLA benchmarks~\cite{zhong2019publaynet,hj2020dataset} opened a new direction for deep-learning-based approaches.  Later, Biswas et.al.~\cite{biswas2021beyond} redefined the DLA task as an instance-level segmentation task to detect both bounding boxes along with segmentation masks, especially in layouts of pages containing overlapped objects. More recently, transformer-based approaches~\cite{biswas2022docsegtr,huang2022layoutlmv3,banerjee2023swindocsegmenter} claimed the recent state-of-the-art on DLA, especially on large-scaled document datasets. However, there is still a huge scope for improvement of transformer methods, especially in smaller annotated datasets like ~\cite{clausner2019icdar2019}. Recently, language-based approaches like LayoutLMv3~\cite{huang2022layoutlmv3} and UDoc~\cite{gu2021unidoc} have been tested on PubLayNet~\cite{zhong2019publaynet} benchmark to get the best performance. But they fail in document benchmarks with more complex layouts and much smaller annotated samples. In this work, we strive to propose SelfDocSeg a pure vision-based self-supervised strategy to mitigate the aforementioned issues. 

% *Biswas et al
% *DocSegTr
% *LayoutParser
% *LayoutLMv3
% *

\section{Methodology}\label{sec:method}
\subsection{Problem Formulation} In the context of document layout segmentation we have access to a relevant dataset $\mathcal{D} = \{x, y\}$ where $x \in \mathcal{I}^{3 \times H \times W}$ is a standard RGB document image with height $H$ and width $W$, and $y = \{y_1, \ldots, y_p\}$ with $y_l$ having a set of coordinates $y_l^{\texttt{mask}}$ of object pixels and $y_l^{\texttt{label}}$ for the object $l = \{1, \ldots, p\}$ assuming $p$ objects in the image $x$. However, according to the core principle of self-supervised learning, we do not use this dataset as it contains the ground-truth annotations, and thus we derive a dataset $\mathcal{D}' = \{x, m\}$ from dataset $\mathcal{D}$ such that it contains only the images without annotated data and a rough binary mask $m$ extracted from $x$ as described in Section~\ref{sec:method-mask} depicting the physical layout of the document image.

The pre-training strategy for the image encoder briefly described in Section~\ref{sec:method-pretrain}, has been adapted to our self-supervised framework, SelfDocSeg. Formally, we train an image encoder $F_\theta$ parameterized by $\theta$ in a self-supervised setting specifically designed to cater to document object recognition. Once the pre-training is done we apply these weights to initialize the backbone of an object detector model to be later fine-tuned for the segmentation of document objects as described in Section~\ref{sec:method-finetune}. 

\begin{figure*}[t]
\centering
\includegraphics[width=\linewidth]{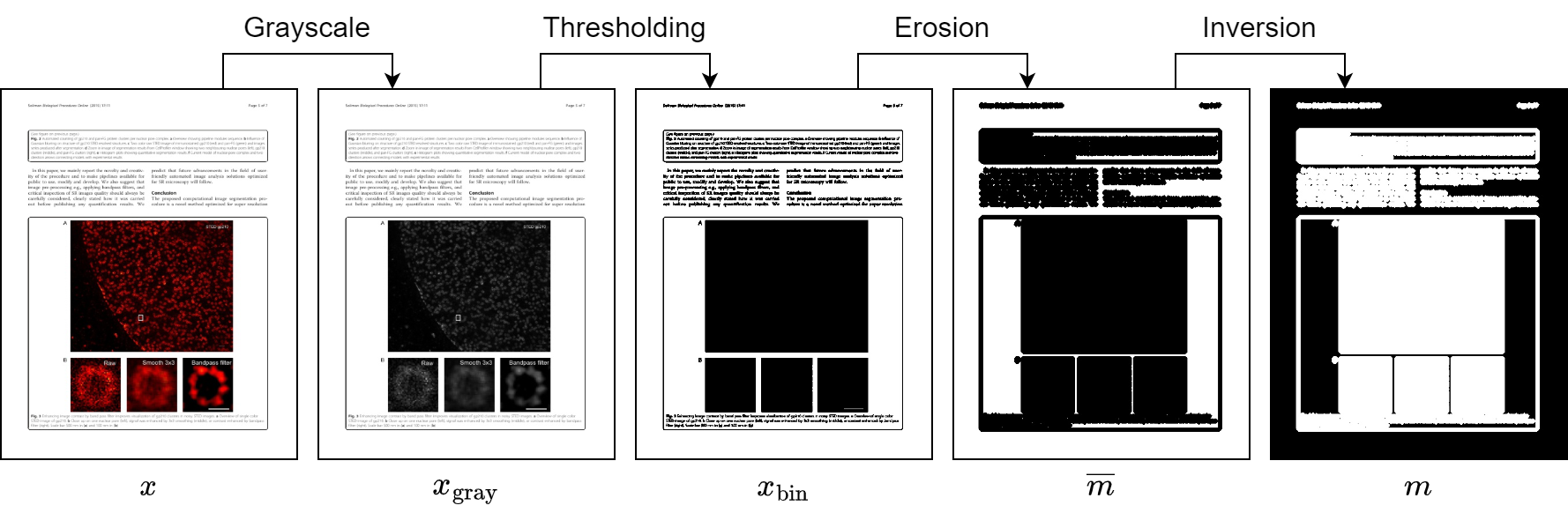}
%\vspace{-0.6cm}
\caption{\textbf{Layout Mask Generation Pipeline}. The figure illustrates a stepwise approach with different strategies adapted for generating the final layout mask for a given document image.}

\label{fig:mask}
%\vspace{-0.4cm}
\end{figure*}

\subsection{Layout Mask Generation}\label{sec:method-mask}

The layout mask generation is a crucial step for the proposed pipeline as our self-supervised framework depends on it for some visual guidance. For any document image $x$ we generate a mask $m$ using classical image processing techniques as depicted in Fig.~\ref{fig:mask}. At first, we convert the RGB image $x$ to grayscale $x_{\text{gray}}$ as defined by CIE, followed by global thresholding to obtain a binarized output $x_{\text{bin}}$. We then perform an erosion operation over $x_{\text{bin}}$ to make the logical layout elements of the document like characters, sub-figures, and plot lines a little thicker and preferably a blob, so that the blobs get connected to form a rough physical layout $\overline{m}$. The final layout mask $m$ is generated by an inversion operation of this eroded rough layout $\overline{m}$.

\subsection{Encoder Pre-Training}\label{sec:method-pretrain}
\paragraph{\textbf{Overview:}} The pre-training of the image encoder is quite difficult to achieve as we are specifically considering the object detection problem, a multi-task challenge that enforces the encoder to learn important salient features contributing towards inter-class variance of different logical layout components and simultaneously localizing the object regions. To overcome this problem, we use the layout masks as visual guidance. However, there is no option to use layout masks directly as we have two distinct yet related tasks at hand. Thus, the challenge and also the scope of innovation remain on how the guidance is provided for the two aforementioned tasks.

\begin{figure*}[t]
\centering
\includegraphics[width=\linewidth]{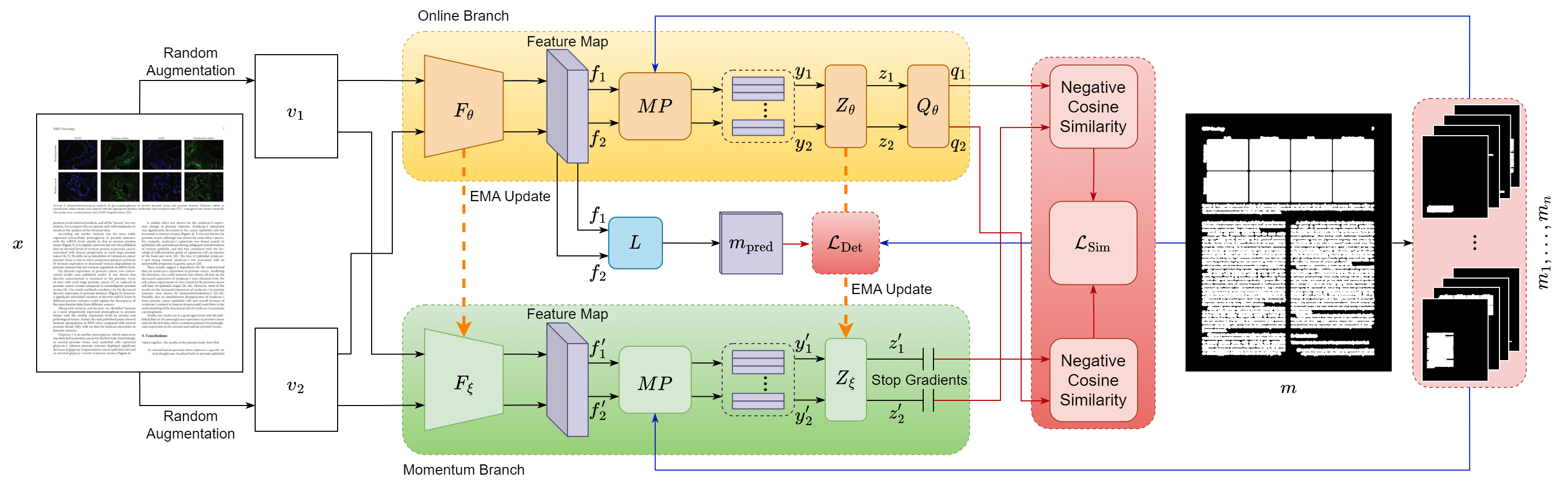}
%\vspace{-0.6cm}
\caption{
\textbf{Model Overview.} A simple architectural design of SelfDocSeg pre-training modeled after the BYOL~\cite{grill2020bootstrap} framework along with a layout prediction module.}

\label{fig:arch}
%\vspace{-0.4cm}
\end{figure*}

The proposed architecture of SelfDocSeg is developed inspired by the popular BYOL~\cite{grill2020bootstrap} self-supervised framework. It mainly consists of two branches, online and momentum, parameterized by two sets of weights $\theta$ and $\xi$ respectively. The overall learning strategy of the architecture is to use a self-distillation technique that involves the two branches of the network which exploits the similarity in salient features extracted from two different views or variations of the same input. The feature similarity exploitation principle demands the generation of significantly different views of the input document image $x$. Thus we augment $x$ randomly to generate two different versions $v_1, v_2$ and feed them to the two branches of the SelfDocSeg framework as illustrated in Fig.~\ref{fig:arch}. Each of these branches employs multiple modules to generate meaningful semantic embeddings from the image through a mask pooling operation. This operation is designed in a way that simultaneously extracts embeddings for all possible layout objects as a batch from the input image $x$ using the guidance of the layout mask $m$ as we roughly extract separate masks of each layout object to obtain an average pooled vector for each.

In parallel, the layout prediction module ensures that the online network learns to locate regions of interest in the document image feature map. We use both the feature maps from $v_1, v_2$ to feed the layout predictor module and generate pixel-wise probability scores of a logical document object being present at that pixel. This whole module is trained using $m$ with the help of focal loss\cite{lin2017focal}. The overall architecture is depicted in Fig.~\ref{fig:arch} and the overall loss function used to optimize the training is given by equation~\ref{equ:loss_total} where $\mathcal{L}_{\text{Det}}$ comes from the \textit{Layout Prediction Module} and $\mathcal{L}_{\text{Sim}}$ comes from \textit{Layout Object Representation Learning} as discussed later in the following subsections.

\begin{equation}
\label{equ:loss_total}
\begin{aligned}
    \mathcal{L}_{\text{total}} = \mathcal{L}_{\text{Det}} + \mathcal{L}_{\text{Sim}}
\end{aligned}
\end{equation}

\paragraph{\textbf{Augmentation:}} The augmentations used to synthesize $v_1, v_2$ from the input document image $x$ are well crafted to match the task and are very similar to the augmentations used in SimCLR~\cite{pmlr-v119-chen20j}. From the set of augmentations used by SimCLR, we carefully exclude random cropping and random horizontal flip. As we are performing mask pooling on the feature map, it simultaneously tries to learn all the possible physical layout object representations. And thus, taking two patches from the image and using them as positive or contrastive pairs is not required and is inherently done for multiple pairs simultaneously. Moreover, the layout predictor module is designed to predict layout all over the document image and thus the task does not align well with taking randomly cropped patches from the document image. On the other hand, the inputs, being document images, contain a large variety of texts and glyphs which could break the clearly established pattern when flipped at random and thus can hinder the training of our module. So, we exclude these two from the set of augmentations and the final set includes the following: (a) Gaussian blurring, (b) color jittering, (c) color dropping, and (d) solarization. The augmentations are not manually controlled and are randomly applied on input image $x$ with random parameters which provide a wide range of variations as required to learn the similarity between the two views $v_1, v_2$.

\paragraph{\textbf{Image Encoder \& Mask Pooling:}} We define the image encoder $F: \mathcal{I}^{3 \times H \times W} \rightarrow f$ which takes an input image and encodes it to a salient feature map $f \in \mathbb{R}^{c \times h \times w}$ with $c$ channels and height and width of $h$ and $w$ respectively. Considering both the online and the momentum branches, we have two image encoders $F_\theta$ and $F_\xi$ providing feature maps $f$ and $f'$ respectively.

The mask pooling operation is performed on feature maps $f$ and $f'$ separately on the two branches to get all the layout object representation vectors simultaneously. The individual object layout masks $m_{k=1,\ldots,n}$ are obtained by detecting individual contours from $m$ assuming $n$ separate contours in $m$. Once we have all the possible layout object masks, we get the average pooled vector $y^{(k)} \in \mathbb{R}^d$ with dimension $d$ for each of the objects from the feature map $f$ according to the established definition of Mask Pooling\cite{henaff2022object} operation $MP: f \rightarrow y$ as per equation~\ref{equ:maskpool} where $i,j$ are pixel coordinates. Once we get $y^{(k)}$ for all $k=1,\ldots,n$ estimated objects from $m_k$ in layout mask $m$, we neatly stack the vectors to form $y \in \mathcal{R}^{k \times d}$.

\begin{equation}
\label{equ:maskpool}
\begin{aligned}
    y^{(k)} = \frac{1}{\sum_{i,j} m_k[i,j]}\sum_{i,j} m_k[i,j]f[i,j]
\end{aligned}
\end{equation}

\paragraph{\textbf{Layout Object Representation Learning:}} The overall framework of SelfDocSeg is designed after BYOL~\cite{grill2020bootstrap} and thus follows a similar strategy to learn the representations from document images using cosine similarity. The online branch comprises an image encoder $F_\theta$, projector module $Z_\theta$, and a predictor module $Q_\theta$ while the momentum branch consists of the image encoder $F_\xi$ and a projector module $Z_\xi$ where $\theta$ and $\xi$ are the sets of parameters for online and momentum branch respectively.

For inputs $v_1, v_2$ we get feature maps $f_1, f_2$ and $f'_1, f'_2$ from $F_\theta$ and $F_\xi$ respectively and we use mask pooling $MP$ as a strategy to get the encoded vector from the feature map which is designed to pool out all the approximate physical layout segments into individual encoded vectors simultaneously as discussed earlier and neatly stack the encodings in a batch resulting in $y_1, y_2$ and $y'_1, y'_2$, which contain average pooled vectors for all $k=1,\ldots,n$ objects stacked in batches. The encoded vectors $y_1, y_2$ and $y'_1, y'_2$ are passed through projector modules $Z_\theta$ and $Z_\xi$ to yield $z_1, z_2$ and $z'_1, z'_2$ respectively. As the online branch of the framework has a predictor layer $Q_\theta$, it takes input $z_1, z_2$ and produces $q_1, q_2$.

\begin{equation}
\label{equ:grad}
\begin{aligned}
    \theta \leftarrow \text{optimizer}\left(\theta, \nabla_\theta \mathcal{L}_{\text{total}}, \eta \right) \\
    \xi \leftarrow \tau \xi + \left(1 - \tau \right) \theta
\end{aligned}
\end{equation}

SelfDocSeg is designed to learn object representations without labels in a self-distillation approach and uses exponentially moving average (EMA) to update the weights of momentum branch $\xi$ from the online branch weights $\theta$. Thus, $F_\xi$ and $Z_\xi$ are updated using EMA from $F_\theta$ and $Z_\theta$ while $F_\theta$ and $Z_\theta$ are updated using back-propagation on the loss function as given in equation~\ref{equ:grad} where $\eta$ is the learning rate, $\tau$ is the momentum of EMA and $\nabla_\theta\mathcal{L}_{\text{total}}$ denotes the gradient corresponding to the total loss $\mathcal{L}_{\text{total}}$ from equation~\ref{equ:loss_total}. It is also to note that the back-propagation does not happen in the momentum branch as denoted by the 'stop gradient' sign in Fig.~\ref{fig:arch}. The similarity loss $\mathcal{L}_{\text{Sim}}$ uses cosine similarity to compare representations from online and momentum branches as per equation~\ref{equ:loss_byol}.

\begin{equation}
\label{equ:loss_byol}
\begin{aligned}
    \mathcal{L}_{\text{Sim}} = 4 - 2 \left(\frac{\langle q_1,z'_2 \rangle}{||q_1||_2 \cdot ||z'_2||_2} + \frac{\langle q_2,z'_1 \rangle}{||q_2||_2 \cdot ||z'_1||_2}\right)
\end{aligned}
\end{equation}

\paragraph{\textbf{Layout Prediction Module:}} The layout prediction module $L$ is an auxiliary module facilitating the layout object localization. This module helps the encoder to learn better representation specifically for the detection task. The module takes input from both the feature maps $f_1, f_2$ and predicts a mask layout $m_{\text{pred}}$, which is then compared to the approximated mask $m$ using focal loss~\cite{lin2017focal}, $\mathcal{L}_{\text{Det}}$ given in equation~\ref{equ:loss_focal} with hyper-parameters $\alpha$ and $\gamma$. The intuition behind this module is to instill a notion of localization of the layout objects, of which the encoder is learning the representation from $\mathcal{L}_{\text{Sim}}$. %The overall loss is given in equation~\ref{equ:loss_total}.

\begin{equation}
\label{equ:loss_focal}
\begin{aligned}
    \mathcal{L}_{\text{Det}} = - \frac{\alpha}{\sum_{i,j} m[i,j]} \cdot \sum_{i,j} (m[i,j](1 - m_{\text{pred}}[i,j])^\gamma \log m_{\text{pred}}[i,j] \\
    + (1 - m[i,j])m_{\text{pred}}[i,j]^\gamma \log (1 - m_{\text{pred}}[i,j]))
\end{aligned}
\end{equation}

%\subsubsection{Representation}

\subsection{Fine-tuning}\label{sec:method-finetune}
Once the image encoder is trained we proceed towards the main task of document segmentation and for this purpose, we take the pre-trained weights of the image encoder $F_\theta$ and use it to initialize the backbone weights of an object detector model. This model is trained with the supervision of ground-truth labels using whole annotated dataset $\mathcal{D}$ and we use Mask RCNN\cite{he2017mask} as our object detector model equipped with a feature-pyramid network (FPN)\cite{lin2017feature} from intermediate layers for multi-scale detection.

\section{Experiments}\label{sec:exp}
% \input{./tex/results.tex}
% \subsubsection{Datasets:} For the training and evaluation of our framework for document segmentation, we used several datasets specifically designed for the task. For the purpose of pre-training the backbone image encoder, we use PubLayNet\cite{} and DocLayNet\cite{} datasets. The PubLayNet dataset contains 335,703 training, 11,245 validation, and 11,405 test images with labeled layout masks of five different layout object classes. The DocLayNet dataset contains 69,375 training, 6,489 validation, and 4,999 test images from six different domains with annotated ground-truth labels for 11 separate classes. For pre-training with both datasets, we use only the training split without the ground-truth labels as mentioned in~\ref{sec:method}.

\paragraph{\textbf{Datasets:}} For the training and evaluation of our framework for document segmentation, we used several datasets specifically designed for the task. We use the DocLayNet\cite{pfitzmann2022doclaynet} dataset for our pre-training experiments. It contains 69,375 training, 6,489 validation, and 4,999 test images from six different domains with annotated ground-truth labels for 11 separate classes. However, we only use the training split without the ground-truth labels for the pre-training phase as mentioned in Section~\ref{sec:method}. 

Once the pre-training is complete we move on to the document layout analysis task to evaluate the efficacy of our pre-training strategy and for the same, we use PRImA\cite{prima2009read} dataset with 305 labeled images in total; Historic Japanese\cite{hj2020dataset} dataset, having 181,097 training, 39,410 validation and 39,109 test layout objects of seven different categories spanning over 2,271 document images; and the extensive PubLayNet~\cite{zhong2019publaynet} dataset containing 335,703 training, 11,245 validation, and 11,405 test images with labeled layout masks of five different layout object classes, along with the DocLayNet~\cite{pfitzmann2022doclaynet} dataset used in pre-training.

\paragraph{\textbf{Implementation Details:}} The complete self-supervised framework is developed using a self-supervised library named Lightly-AI~\cite{susmelj2020lightly}, that is built on PyTorch Lightning~\cite{falcon2019pytorch} and PyTorch~\cite{NEURIPS2019_9015}. All the models are trained on a 48 GB Nvidia RTX A40 GPU.

The mask generation process using classical image processing, described in \ref{sec:method-mask}, is developed using OpenCV~\cite{opencv_library}. We empirically found the global threshold value of $239$ working well enough for the pre-training dataset, DocLayNet~\cite{pfitzmann2022doclaynet} given that the grayscale images are having 8-bit unsigned integer datatype. For the erosion operation, we use a $5 \times 5$ rectangular kernel as the structuring element.

%The Mask RCNN\cite{he2017mask} is used as an object detector and is built from Detectron 2~\cite{wu2019detectron2} with PyTorch.

Firstly, for the image encoders $F_\theta, F_\xi$ we use a standard ResNet50~\cite{he2016deep} and we use the last residual block for extracting encoded feature maps, leaving out the global average pooling and fully-connected layers at the end, and thus the average pooled object vectors have dimension $d=2048$ same as $c$ in the output of last residual block. The projectors $Z_\theta$ and $Z_\xi$, and the predictor $Q_\theta$, are implemented as two-layer multi-layer perceptrons (MLPs) with $4096$ hidden and $256$ output dimensions. The auxiliary layout prediction module $L$ is implemented by a $1 \times 1$ convolution block to predict the layout mask. The focal loss $\mathcal{L}_{\text{Det}}$ mentioned in equation~\ref{equ:loss_focal} uses two hyper-parameters, a weighing factor $\alpha=0.25$ and a focusing parameter $\gamma=2$.

In the pre-training phase, the optimal model parameters are learned by optimizing the loss $\mathcal{L}_{\text{total}}$ from equation~\ref{equ:loss_total} using LARS~\cite{you2017large} optimizer with an initial learning rate of $\eta=0.2$, and a weight decay of $0.0005$. The learning rate is decayed following a cosine annealing schedule~\cite{loshchilov2016sgdr} for $800$ epochs. For the momentum branch, the value of the momentum hyper-parameter $\tau$ is set to $0.99$.

The object detection model in fine-tuning with task-specific ground-truth annotation uses the same backbone, ResNet50~\cite{he2016deep} for Mask RCNN~\cite{he2017mask} with FPN~\cite{lin2017feature} and is trained with Detectron2~\cite{wu2019detectron2}. We use this framework because it is built on top of PyTorch, is extremely easy to use, and is reliable for deploying models. We have used Nvidia RTX A40 GPUs for all of our fine-tuning purposes. Our trained models have been based on ResNet-50 weights pre-trained as per our proposed framework, SelfDocSeg, on the DocLayNet~\cite{pfitzmann2022doclaynet} dataset. %We train the layers with our dataset after initializing the weights on the already trained models.
The initial learning rate of the model was 0.0025, and it is trained for a total of 300,000 iterations. A multitude of anchor scales and aspect ratios are considered to cover the input image, resulting in the number of anchor boxes $k=64$. Nesterov Momentum was used to train with the Stochastic Gradient Descent (SGD) optimizer \cite{dozat2016incorporating}, and the batch size per image in the RoI heads was 128. Our task had a detection non-maximum suppression (NMS) threshold of $0.4$ and a detection minimum confidence score of $0.6$. The data loader's workforce was set at four. We set the model's testing threshold in the RoI heads to $0.6$ following the completion of the fine-tuning process. The values mentioned are carefully chosen after reviewing empirical data.

\subsection{Competitors}\label{sec:exp-sota}

\paragraph{\textbf{State-of-the-Art:}} In this work, our contribution is towards a novel self-supervised pre-training strategy, which cannot be directly compared to its counterparts without having to look at the performance of the downstream task. Thus, we compare the methodologies with existing models after fine-tuning the downstream task using a document object detector.

In the context of self-supervised DLA task, the research community has yet to see various strategies and thus we have only a few strategies to compare with. We compared our methods with state-of-the-art (SOTA) self-supervised methodologies: (a) LayoutLMv3~\cite{huang2022layoutlmv3} uses a multi-modal transformer that uses a masked language modeling and masked image modeling-based strategy using linear image patch embeddings from the document and textual embeddings extracted from a pre-trined optical character recognition (OCR) and also tries to align word patches with image patches jointly with an image and word token classification task. (b) UDoc~\cite{gu2021unidoc} employs a region of interest alignment strategy that aligns the visual features from the document images with the textual embeddings and location information from a pre-trained OCR to generate joint embeddings at the sentence level to use them with contextual masking  to learn with visual contrastive learning and visual-textual alignment through a cross attention module. The noticeable difference between these existing SOTA with the proposed pipeline can be identified as the use of pre-trained OCR to guide the pre-training with textual and layout cues along with visual features while we use only visual features for the purpose. (c) DiT~\cite{li2022dit} uses only visual features like the proposed strategy using masked image modeling with BEiT~\cite{bao2022beit} pre-training over a massive dataset of size 42 million approximately.

In parallel, we show a comparative analysis with supervised SOTA methodologies: (a) DocSegTr~\cite{biswas2022docsegtr} implements a transformer encoder-decoder architecture, being the first of its kind, along with a convolutional backbone for document segmentation. (b) LayoutParser~\cite{shen2021layoutparser} uses a universal framework built on top of CNN-based approaches for DOD evaluation (c) Biswas \etal~\cite{biswas2021beyond} had used a Mask RCNN-styled architecture employing a fully convolutional network, for instance, level segmentation and class prediction and bounding box regression empowered by a region proposal network to suggest significant regions of interest from the features extracted at multiscale by FPN on a backbone.

\begin{table}[t]
\centering
\caption{Quantitative analysis of the performance of the document object detection task along with the guidances used during supervision. V, L, and T stand for visual, layout, and textual cues respectively.}\label{tab:performance}
\begin{tabular}{lccccccccc}
\toprule
\multirow{2}{*}{} & \multirow{2}{*}{Methods} & \multicolumn{3}{c}{Cues} & \multirow{2}{*}{\# Data} & DocLayNet & PubLayNet & PRImA & HJ\\
                           & & V & L & T & & mAP & mAP & mAP & mAP\\
\hline
\multirow{4}{*}{Supervised} & DocSegTr~\cite{biswas2022docsegtr} & \cmark & \xmark & \xmark & - & - & 90.4 & 42.5 & \textbf{83.1}\\
                    % & LayoutLMv3~\cite{huang2022layoutlmv3} & \cmark & \cmark & \cmark & - & 95.1 & 40.3 & 82.7\\
                    % & UDoc~\cite{gu2021unidoc} & \cmark & \cmark & \cmark & - &  93.9 & - & -\\
                    % & DocSegTr~\cite{biswas2022docsegtr} & \cmark & \xmark & \xmark & - & 90.4 & 42.5 & 83.1\\
                    & LayoutParser~\cite{shen2021layoutparser} & \cmark & \cmark & \cmark & - & - & 86.7 & \textbf{64.7} & 81.6\\
                    & Biswas \etal~\cite{biswas2021beyond} & \cmark & \xmark & \xmark & - & - & 89.3 & 56.2 & 82.0\\
                    & Mask RCNN~\cite{he2017mask} & \cmark & \xmark & \xmark & - & 72.4 & 88.6 & 56.3 & 80.1\\
\hline
% \multirow{3}{*}{Baselines} &  BYOL+M RCNN & - & - & - & -\\
\multirow{2}{*}{\makecell{\\Self-\\Supervised}} & LayoutLMv3$_\text{Base}$~\cite{huang2022layoutlmv3} & \cmark & \cmark & \cmark & 11M & - & \textbf{95.1} & 40.3 & 82.7\\
                                 & UDoc~\cite{gu2021unidoc} & \cmark & \cmark & \cmark & 1M & - &  93.9 & - & -\\
                                 & DiT$_\text{Base}$~\cite{li2022dit} & \cmark & \xmark & \xmark & 42M & - &  93.5 & - & -\\
                            % & BYOL + M RCNN & \cmark & \xmark & \xmark & 63.5 & 79.0 & 28.7 & 59.8\\
                            % & M RCNN (Imagenet Pre-trained) & - & - & - & -\\
                            % & M RCNN (Random Initialization) & - & - & - & -\\
                            % & M RCNN (Fully Supervised) & \cmark & \xmark & \xmark & 72.4 & 88.6 & 56.3 & 80.1\\
%4th-level heading & {\itshape Lowest Level Heading.} Text follows & 10 point, italic\\
\hline
\multirow{2}{*}{Proposed} & BYOL~\cite{grill2020bootstrap} & \cmark & \xmark & \xmark & 81k & 63.5 & 79.0 & 28.7 & 59.8\\
                          & \textbf{SelfDocSeg} & \cmark & \xmark & \xmark & 81k & \textbf{74.3} & 89.2 & 52.1 & 78.8\\
\bottomrule
\end{tabular}
\end{table}

\paragraph{\textbf{Comparable Baselines:}} In terms of DOD, we also establish a few baselines, both supervised and unsupervised. We used BYOL~\cite{grill2020bootstrap} as a pre-training strategy to train an image encoder and used it to perform the fine-tune training on the document object detector, Mask RCNN~\cite{he2017mask} with ground-truth and consider it as our potential self-supervised baseline. On the other hand, we train Mask RCNN~\cite{he2017mask} in a fully supervised setting with ground-truth annotations. %both from Imagenet~\cite{5206848} pre-trained initialization and random initialization for the supervised baselines.

\begin{figure*}[t]
%\captionsetup[subfloat]{farskip=0pt,captionskip=0pt}
\subfloat[Invoice]{%
  \includegraphics[width=0.25\linewidth,height=4cm]{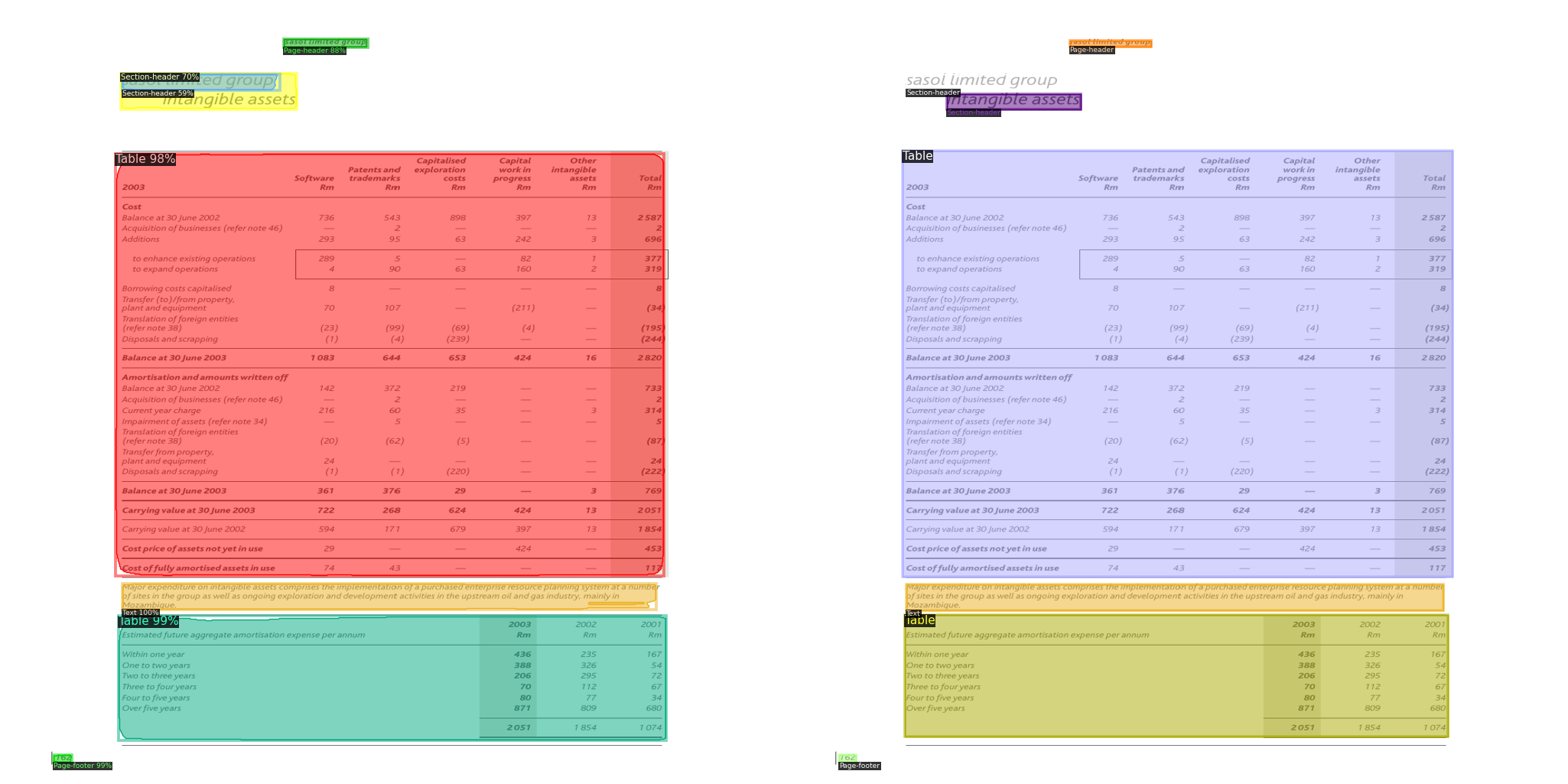}%
}
\subfloat[Advertisement]{%
  \includegraphics[width=0.25\linewidth,height=4cm]{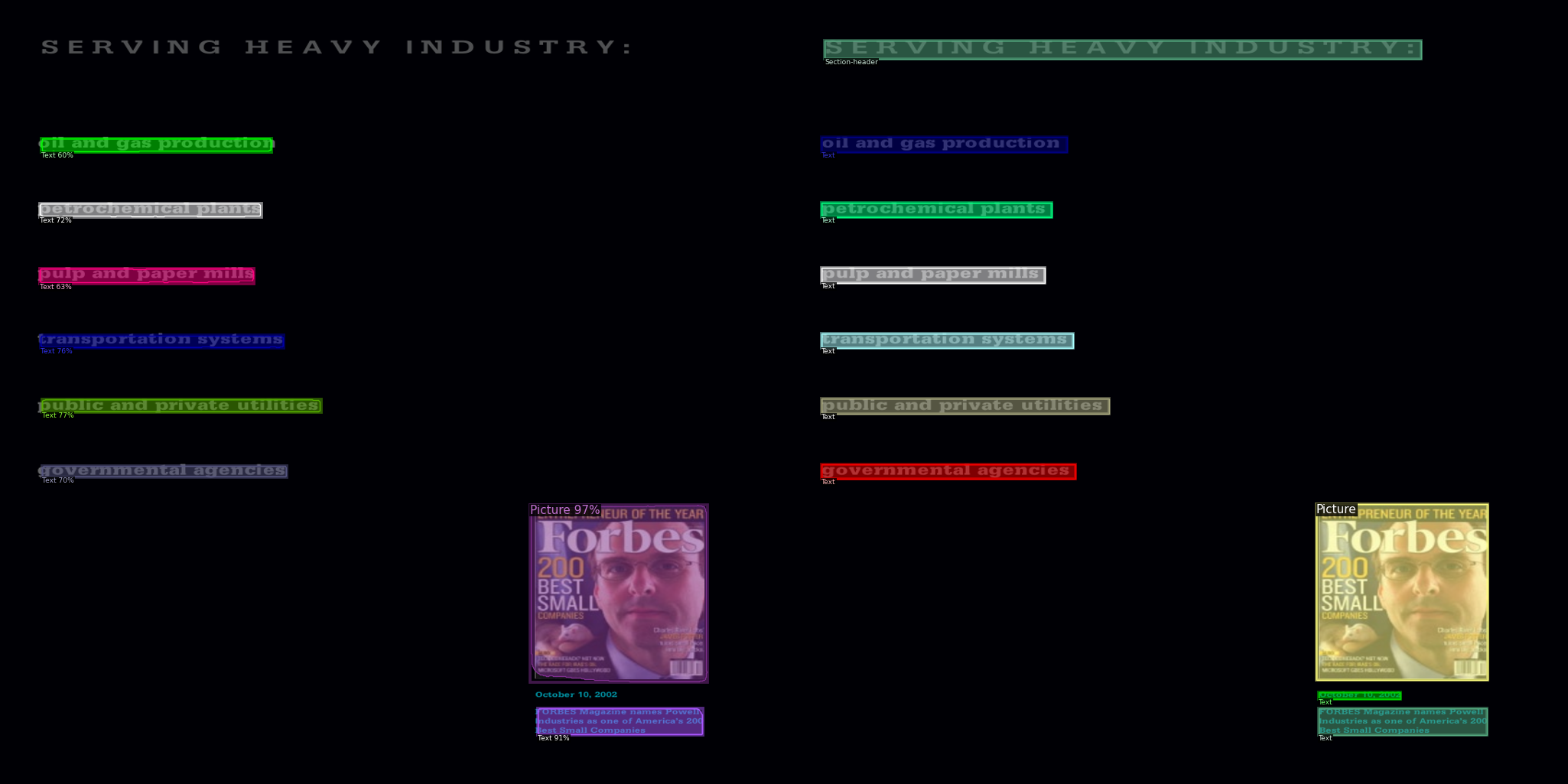}%
}
\subfloat[Industrial]{%
  \includegraphics[width=0.25\linewidth,height=4cm]{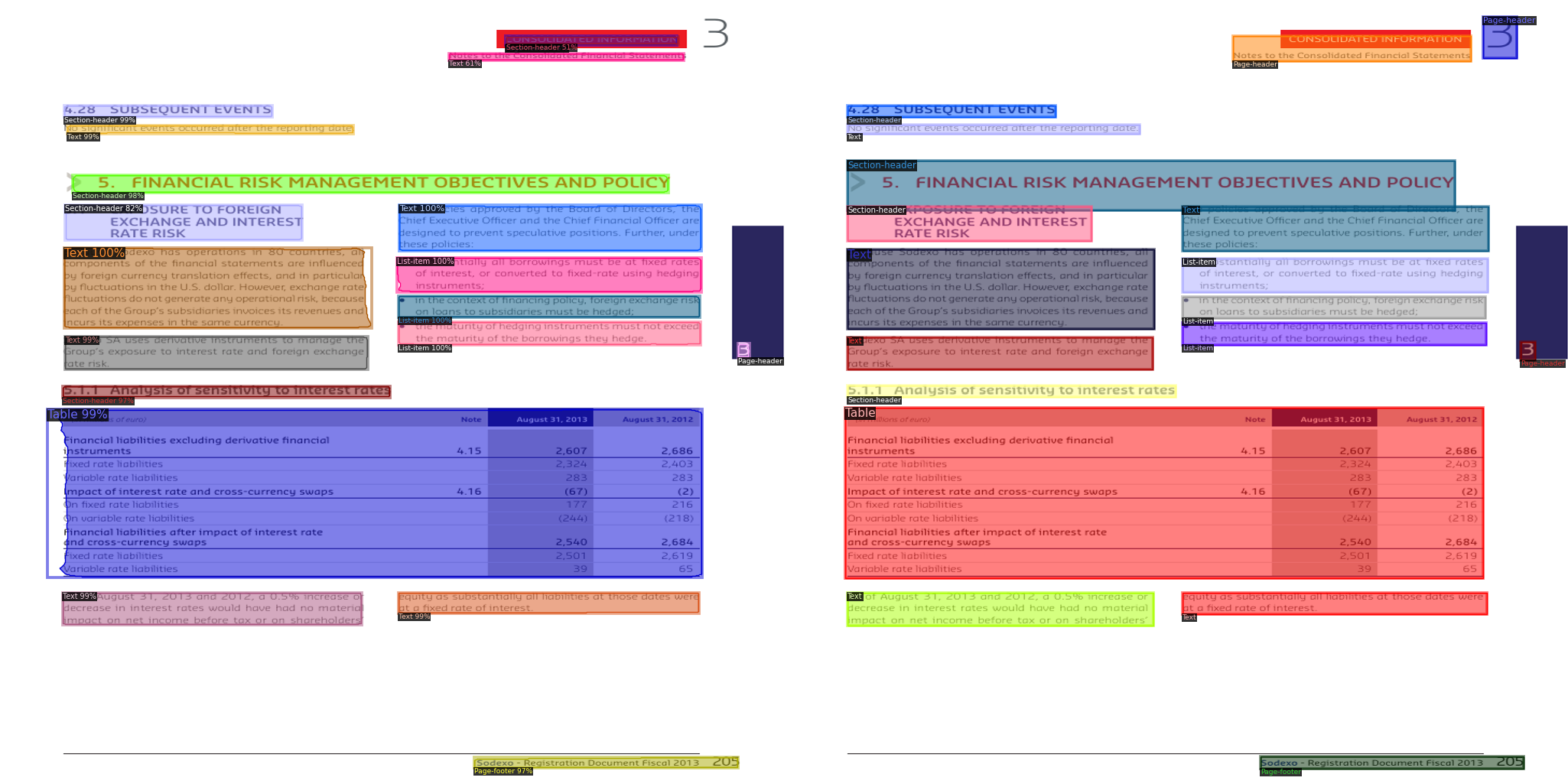}
}
\subfloat[Leaflet]{%
  \includegraphics[width=0.25\linewidth,height=4cm]{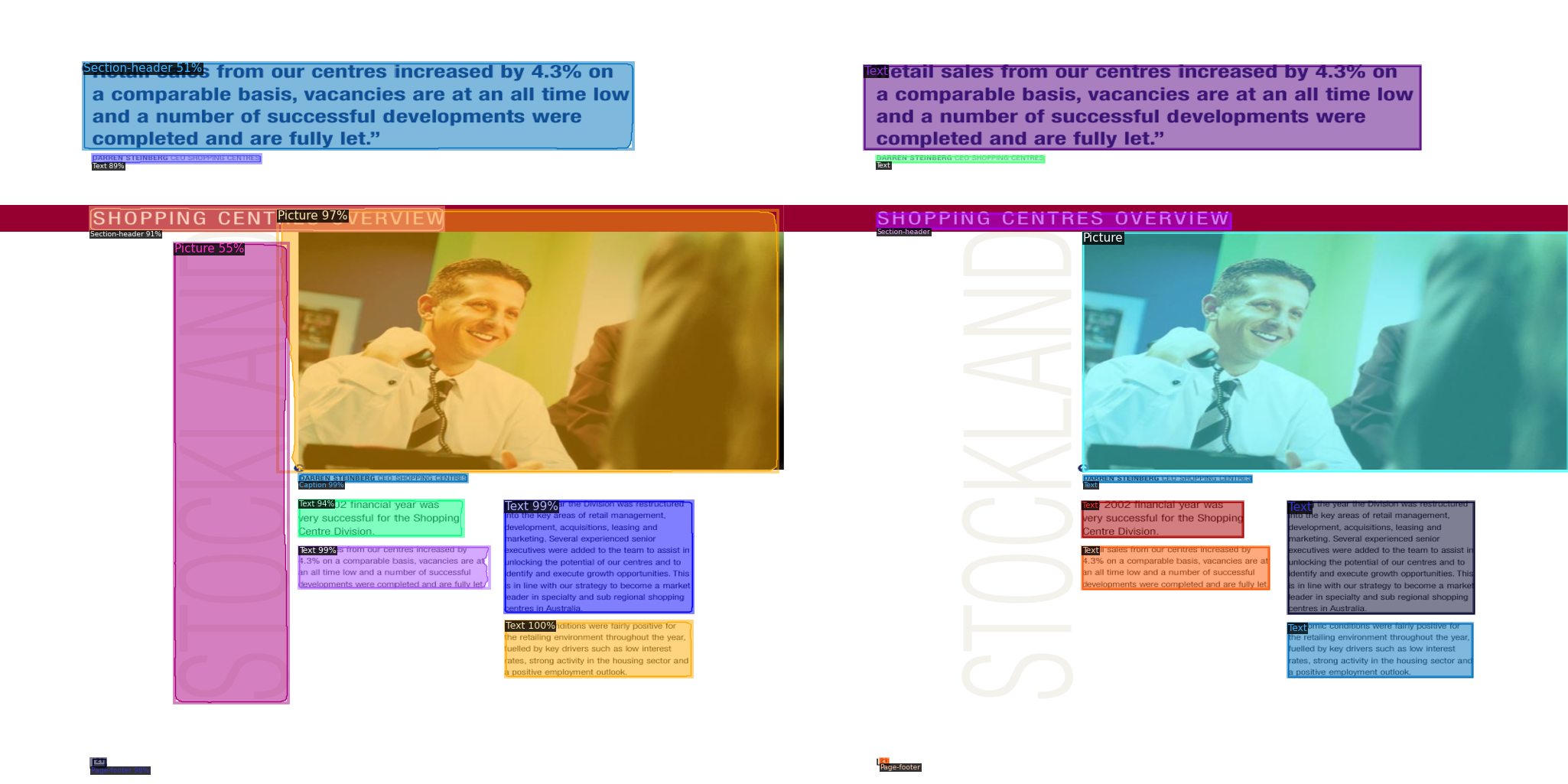}
}
\caption{Qualitative analysis of the SelfDocSeg framework on the DocLayNet datasets (\textbf{Left:} Predicted layout \textbf{Right:} Ground-truth)}
\label{fig:examples_detection_all}
\end{figure*}

\subsection{Performance Analysis}\label{sec:exp-performance}
The performance of our pre-training strategy is compared with the existing methodologies in Table~\ref{tab:performance} in terms of mean average precision (mAP) score at the pixel level segmentation of document objects along with the depiction of guidance/cues used for each of the methods and amount of data used for self-supervised pre-training. (a) LayoutLMv3~\cite{huang2022layoutlmv3} and UDoc~\cite{gu2021unidoc} have superior performance on PubLayNet~\cite{zhong2019publaynet} dataset being the current self-supervised SOTAs. LayoutLMv3~\cite{huang2022layoutlmv3} also has superior performance over other works. However, both methods use pre-trained OCR for textual and layout cues which provides them an additional advantage in learning instance-level document layout component embeddings coupled with the superior backbone of a multi-modal transformer. In comparison, the proposed SelfDocSeg performs at par without any guidance from pre-trained OCR. (b) DiT$_\text{Base}$~\cite{li2022dit} has a superior performance compared to the proposed strategy due to its powerful Vision Transformer~\cite{dosovitskiy2020image} backbone coupled with the huge pre-training dataset of 42 million. (c) DocSegTr~\cite{biswas2022docsegtr} uses a convolutional backbone as well as a transformer architecture with multi-headed attention that is inherently more powerful for extraction of useful features and thus supervision with ground truth provides superior performance in this scenario. It is noticed that transformer-based models could not perform well in small datasets like PRImA~\cite{prima2009read}. (d) LayoutParser~\cite{shen2021layoutparser} is a comprehensive layout analysis toolkit that offers state-of-the-art performance with common CNN-based approaches, which shows decent results owing to a powerful OCR module along with document image augmentations. (e) The work of Biswas \etal~\cite{biswas2021beyond} employs a modified Mask RCNN architecture with end-to-end supervision and performs decently in the context of Mask RCNN~\cite{he2017mask}, compared to vanilla Mask RCNN in the fully supervised setting. (f) The vanilla BYOL pre-training fails to learn meaningful document object representations and thus performs poorly in Mask RCNN. The possible justification for the performance could be attributed to unrestricted random cropping and flipping in augmentations, which leads to two different classes present in the same input document image being treated as positive pairs. (g) Vanilla Mask RCNN performs decently in the fully supervised setting. (h) It is evident that although Mask RCNN does not have all-over superiority in performance, it easily outperforms all the transformer-based models in small datasets like PRImA~\cite{prima2009read}. (i) It is also noticed that modern self-supervised strategies use a large amount of data for pre-training. In comparison, the proposed framework has achieved similar performance with significantly less data in pre-training.

To this end, the SelfDocSeg framework proves to be performing at par with the Mask RCNN competitors if not outperforming. And at the same time, it establishes that self-supervised pre-training with only visual features using a limited amount of data is efficient and effective for document segmentation tasks. % OR may be superior
Some visual results of our proposed pipeline are depicted in Fig. \ref{fig:examples_detection_all}.

% \begin{figure*}[t]
% %\captionsetup[subfloat]{farskip=0pt,captionskip=0pt}
% \subfloat[Invoice]{%
%   \includegraphics[width=0.25\linewidth,height=4cm]{figs/0acbded35d0e0e738aa160c1bffceafb8d43c4bc382aa47222ff8c79d26c8e88.png}%
% }
% \subfloat[Advertisement]{%
%   \includegraphics[width=0.25\linewidth,height=4cm]{figs/0b94f9dc5043a14acb186493975ab5b2284115e64b663831454400fa56c95c6f.png}%
% }
% \subfloat[Industrial]{%
%   \includegraphics[width=0.25\linewidth,height=4cm]{figs/0baad9a56f3e6943480477750434a45fc4f4c531c79e8271d63954a897c1c9bf.png}
% }
% \subfloat[Leaflet]{%
%   \includegraphics[width=0.25\linewidth,height=4cm]{figs/1e08caa7b9508fdb5cc49c80b3950eecc594e8f35049ca5818420bf9a86f6580.png}
% }
% \caption{Qualitative analysis of the SelfDocSeg framework on the DocLayNet datasets (\textbf{Left:} Predicted layout \textbf{Right:} Ground-truth)}
% \label{fig:examples_detection_all}
% \end{figure*}

% \begin{table}[t]
% \centering
% \caption{Details of Mask RCNN fine-tuning with variations of training data used.}\label{tab:ablation_percentage_data}
% \begin{tabular}{|l|l|}
% \hline
% Percentage of ground-truth annotations used & mAP on DocLayNet\\
% \hline
% 10\% & -\\
% 50\% & -\\
% 100\% (proposed) & -\\
% \hline
% \end{tabular}
% \end{table}

\begin{table}[t]
\centering
\caption{\textbf{Semi-supervised scenario}. We evaluate the performance of Mask RCNN fine-tuning with varying \% of labeled data used in DocLayNet.}\label{tab:ablation_percentage_data}
\begin{tabular}{ccc}
\toprule
% Percentage of ground-truth annotations used & mAP on DocLayNet\\
\% Annotations & mAP \\
\hline
10\% & 41.3\\
50\% & 65.1\\
100\% & 74.3\\
\bottomrule
\end{tabular}
\end{table}

\begin{table}[t]
\centering
\caption{\textbf{Ablation Study}. Contribution of individual learning objectives during SelfDocSeg pre-training on DocLayNet.}\label{tab:ablation_loss}
\begin{tabular}{cc}
\toprule
Loss & mAP \\
\hline
w/o $\mathcal{L}_{\text{Sim}}$ & 39.1\\
w/o $\mathcal{L}_{\text{Det}}$ & 69.7\\
Combined ($\mathcal{L}_{\text{total}}$) & 74.3\\
\bottomrule
\end{tabular}
\end{table}

\subsection{Ablation Study}\label{sec:exp-ablation}
\paragraph{\textbf{How effective and generalizable is SelfDocSeg pretraining?}} The importance of pre-training can be realized in the performance of the fine-tuned document segmentation model and the natural question lingers around the efficacy of the SelfDocSeg pre-training strategy. To answer the same, we perform several experiments on the fine-tuning stage in a semi-supervised setting. Meaning, we use multiple smaller sets of annotated training data in fine-tuning Mask RCNN and record the performance in Table~\ref{tab:ablation_percentage_data}. Although using all the available training data has the best result, it is evident that the model generalizes well enough due to the learning of extensive and useful visual feature extraction during pre-training as decreasing the amount of labeled data during fine-tuning does not seem to affect the model accuracy much. We see the mAP value drops 9\% only while annotated training data is reduced to half. However, it starts getting affected as the quantity of labeled data drops drastically as we see more than 30\% drop when the annotated dataset is reduced to 10\%.

% \begin{table}[t]
% \centering
% \caption{\textbf{Ablation Study}. Contribution of individual learning objectives during pre-training.}\label{tab:ablation_loss}
% \begin{tabular}{cc}
% \toprule
% Loss & mAP on DocLayNet\\
% \hline
% w/o $\mathcal{L}_{\text{Sim}}$ & 39.1\\
% w/o $\mathcal{L}_{\text{Det}}$ & 69.7\\
% Combined ($\mathcal{L}_{\text{total}}$) & 74.3\\
% \bottomrule
% \end{tabular}
% \end{table}

\paragraph{\textbf{How well do the individual losses contribute in SelfDocSeg?}} We use two loss functions for the proposed pre-training strategy, focal loss in equation~\ref{equ:loss_focal} and a representation loss with cosine similarity in equation~\ref{equ:loss_byol}. To dissect how much these losses individually contribute towards the pre-training, we train two separate encoders from scratch, one with focal loss only and another with representation loss only. The outcomes are depicted in Table~\ref{tab:ablation_loss}. It is evident that the focal loss although helps in localization, it alone is not capable of learning meaningful object representations from the document images. On the other hand, the representation loss is capable enough to learn document object embeddings, but it falls short in the case of the layout analysis tasks. And both the losses together help to pre-train an image encoder for the document segmentation task.

\section{Conclusion}\label{sec:conclusion}
While self-supervision is fairly new in the documents research community, in this paper, we studied the exploitation of rich visual features present in the document images using self-supervision. To this end, we designed a self-supervised strategy to pre-train the image encoder in the document layout analysis context. Our extensive experiments show that the proposed strategy performs decently and generalizes well enough in document images despite having no textual or layout guidance from pre-trained text recognition models. In a word, we present a complete vision-based self-supervised approach towards document segmentation which involves two specific strategies, \ie a pseudo physical layout guided technique for document object localization and a document object representation learning in a self-supervised setting. We further intend to explore the scope of performance improvements in DLA using self-supervision.

\section*{Acknowledgment}
This work has been partially supported by IDEAS - Technology Innovation Hub, Indian Statistical Institute, Kolkata; the Spanish project PID2021-126808OB-I00, the Catalan project 2021 SGR 01559 and the PhD Scholarship from AGAUR (2021FIB-10010). The Computer Vision Center is part of the CERCA Program / Generalitat de Catalunya.
\bibliographystyle{splncs04}
\bibliography{main}




% LNCS Homepage, \url{http://www.springer.com/lncs}. Last accessed 4
% \end{thebibliography}
\end{document}